\documentclass[conference]{IEEEtran}
\IEEEoverridecommandlockouts
\usepackage{cite}
\usepackage{amsmath,amssymb,amsfonts}
\usepackage{algorithmic}
\usepackage{graphicx}
\usepackage{textcomp}
\usepackage{xcolor}
\usepackage{booktabs}
\usepackage{url}
\usepackage{listings}
\usepackage{tabularx}
\usepackage{svg}
\usepackage{graphicx}
\def\BibTeX{{\rm B\kern-.05em{\sc i\kern-.025em b}\kern-.08em
    T\kern-.1667em\lower.7ex\hbox{E}\kern-.125emX}}
\begin{document}

\title{SIC3D: Style Image Conditioned Text-to-3D Gaussian Splatting Generation}



\author{
\IEEEauthorblockN{Ming He, Zhixiang Chen, Steve Maddock}
\IEEEauthorblockA{\textit{
School of Computer Science, University of Sheffield}}\\
\IEEEauthorblockN{\textit{\{mhe27, zhixiang.chen, s.maddock\}@sheffield.ac.uk
}
}
}
\maketitle

\begin{abstract}
Recent progress in text-to-3D object generation enables the synthesis of detailed geometry from text input by leveraging 2D diffusion models and differentiable 3D representations. However, the approaches often suffer from limited controllability and texture ambiguity due to the limitation of the text modality. To address this, we present SIC3D, a controllable image-conditioned text-to-3D generation pipeline with 3D Gaussian Splatting (3DGS). There are two stages in SIC3D.  The first stage generates the 3D object content from text with a text-to-3DGS generation model. The second stage transfers style from a reference image to the 3DGS. Within this stylization stage, we introduce a novel Variational Stylized Score Distillation (VSSD) loss to effectively capture both global and local texture patterns while mitigating conflicts between geometry and appearance. A scaling regularization is further applied to prevent the emergence of artifacts and preserve the pattern from the style image. Extensive experiments demonstrate that SIC3D enhances geometric fidelity and style adherence, outperforming prior approaches in both qualitative and quantitative evaluations. 

\end{abstract}    


\section{Introduction}

Recent advances in large-scale text-to-image diffusion models have substantially accelerated progress in text-to-3D generation. Early methods such as DreamFusion~\cite{poole2022dreamfusion} demonstrated that pre-trained diffusion priors can be used to optimize 3D representations via Score Distillation Sampling (SDS), while Magic3D~\cite{lin2023magic3d} further improved geometric fidelity through a coarse-to-fine optimization strategy. With the emergence of efficient explicit representations, DreamGaussian~\cite{tang2023dreamgaussian} showed that 3D Gaussian Splatting (3DGS) can significantly reduce optimization time while maintaining high-quality reconstruction. More recently, transformer-based approaches such as TRELLIS~\cite{trellis2024} have introduced stronger multi-view reasoning capabilities that enhance geometric coherence. Despite these advancements, existing text-to-3D pipelines primarily aim to improve geometry and consistency, leaving image-conditioned style control largely unexplored.

\begin{figure}[tbh]
    \centering
    \includegraphics[width=0.75\linewidth]{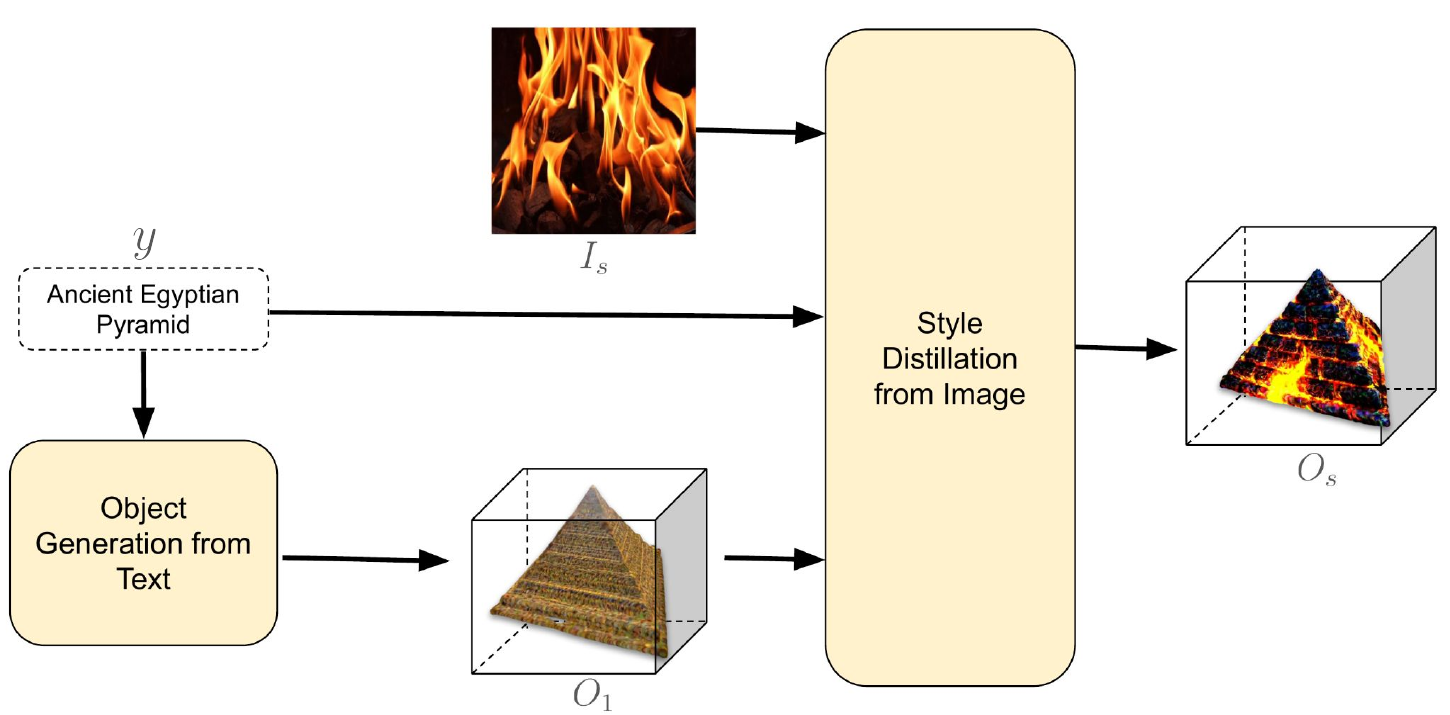}
    \caption{
    \textbf{Overview of the proposed SIC3D framework.}
    The pipeline consists of two stages. 
    In the \emph{Object Generation Stage}, Variational Score Distillation (VSD) is employed to produce a geometrically consistent 3D Gaussian Splatting representation from the text input. 
    In the \emph{Style Distillation Stage}, we introduce Variational Stylized Score Distillation (VSSD), which inject style features from the reference image into the diffusion model. 
    }
    \label{fig:two-stage}
\end{figure}

Beyond text-only control, several works enhance 3D generation with identity or structural constraints, such as DreamBooth3D~\cite{dreambooth3d} and ControlNet-based approaches~\cite{zhang2023adding}. These methods improve controllability but do not address visual style transfer and often rely on slow NeRF-based optimization.

While these control mechanisms broaden the flexibility of 3D generation, they do not address the challenge of transferring visual style, which has been explored primarily in NeRF-based stylization. Dream-in-Style~\cite{kompanowski2024dream} and StyleMe3D~\cite{zhuang2025styleme3d} apply attention-based guidance and Stylized Score Distillation to transfer artistic patterns to NeRF objects, but their optimization is slow and can lead to multi-view inconsistencies. 

Beyond NeRF, Style3D~\cite{DBLP:journals/corr/abs-2412-03571} improves multi-view consistency through image-space stylization followed by reconstruction, while LRM-based stylization~\cite{DBLP:journals/corr/abs-2504-21836} injects style features into a large reconstruction backbone for fast inference. However, both approaches depend on post-hoc reconstruction and do not operate directly on 3D Gaussian primitives, limiting geometry fidelity and fine-grained control. 

Overall, NeRF- and reconstruction-based stylization methods struggle to balance strong style transfer with geometric stability and lack compatibility with 3D Gaussian Splatting, motivating the need for an efficient and controllable 3DGS stylization framework.


To address these challenges, we introduce \textit{SIC3D}, a two-stage framework for efficient and controllable stylization within 3D Gaussian Splatting. By decoupling geometry construction from appearance refinement, SIC3D avoids the entanglement that causes prior stylization methods to suffer from geometric distortion and inconsistent texture guidance. In the first stage, Variational Score Distillation (VSD) is used to obtain a stable, view-consistent 3DGS representation from text. In the second stage, \textit{Variational Stylized Score Distillation (VSSD)} injects image style information through an IP-Adapter–based conditioning mechanism, enabling consistent multi-view stylization while preserving geometry. A lightweight regularization further reduces artifact formation, yielding clean and coherent stylized surfaces.

Our work makes three primary contributions. 
First, we introduce SIC3D, the first image-conditioned stylization framework for 3D Gaussian Splatting, enabling efficient and controllable stylized text-to-3D generation. 
Second, we propose VSSD, which extends VSD with a style injection mechanism that delivers stable optimization and multi-view--consistent stylization. 
Third, we adopt a gradient-based regularization strategy that reduces artifacts arising from geometry--texture inconsistencies, resulting in robust stylization results.

\section{Methodology}

\begin{figure*}[tbh]
    \centering
    \includegraphics[width=0.75\linewidth]{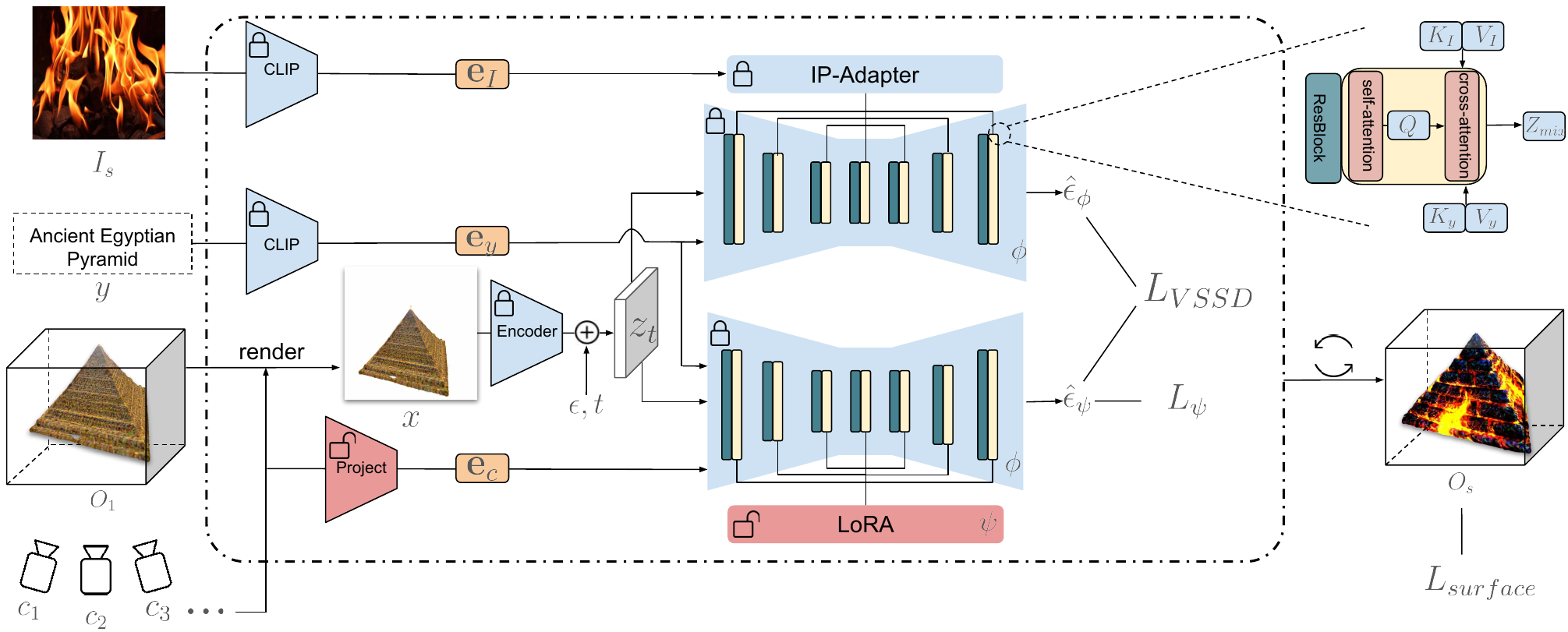}
    \caption{\textbf{Overview of the Style Distillation Stage.} 
    Starting from the first-stage result \(O_1\), generated from text prompt \(y\), two pre-trained diffusion models \(\phi\) augmented with IP-Adapter, LoRA and a camera projection layer are used to obtain the stylized object \(O_s\). 
    The style image \(I_s\) and text \(y\) are encoded by a CLIP encoder into \(e_I\) and \(e_y\). 
    IP-Adapter further processes \(e_I\) and injects style features into the diffusion model via cross-attention, while camera parameters are projected into \(e_c\) to condition the model on viewpoint. 
    During optimization, only the 3DGS parameters, LoRA \(\psi\), and the camera projection layer are updated using the VSSD loss \(L_{\mathrm{VSSD}}\), the scaling loss \(L_{\mathrm{surface}}\), and the diffusion loss \(L_{\psi}\).
    }
    \label{fig:sic3d}
\end{figure*}

In this section, we introduce SIC3D, a two-stage framework for stylized text-to-3D asset generation that distills style features from an input image and applies them to a 3D Gaussian Splatting representation (see Figure~\ref{fig:two-stage}). 

Updating geometry and style simultaneously often couples layout and appearance, leading to unstable optimization and inconsistent guidance across different views. 
To obtain view-consistent stylization, we first generate a geometry-consistent base object \(O_1\) in the \emph{Object Generation Stage}, and then refine its surface appearance with style conditions in the \emph{Style Distillation Stage}. 
Section~\ref{sec:stage1} describes Stage~1, Section~\ref{sec:stage2} introduces Stage~2, and Section~\ref{sec:discussion} discusses particular components of Stage~2. 

\subsection{Stage 1: Object Generation from Text}
\label{sec:stage1}

Recent advances in 3D Gaussian Splatting (3DGS) generation~\cite{yi2023gaussiandreamer, tang2023dreamgaussian, chen2024text} enable high-quality 3D object synthesis from text prompts by iteratively refining a 3D representation using a pre-trained 2D diffusion model. 
Typical pipelines initialize a coarse 3D object---either simple primitives or shapes predicted by 3D diffusion models~\cite{nichol2022point, jun2023shap, lrm}---and update it via diffusion-based guidance.

Score Distillation Sampling (SDS)~\cite{poole2022dreamfusion} is a standard objective that updates the 3D representation by penalizing the discrepancy between the predicted noise \(\hat{\epsilon}_\phi\) and the ground-truth noise \(\epsilon\):
\begin{small}
\begin{equation}
    L_{\mathrm{SDS}} = \mathbb{E}\left[ \omega(t) \, \|\hat{\epsilon}_\phi(z_t; y, t) - \epsilon\|^2 \right],
\end{equation}
\end{small}
\noindent
where \(z_t\) is the perturbed latent of a multi-view rendering \(x = r(O, c)\) under camera \(c\), \(y\) is the text prompt, and \(\omega(t)\) is a time-dependent weight. 
While SDS directly updates the object parameters \(O\), it often yields noisy and inconsistent gradients, resulting in over-smoothed geometry.

To improve fidelity, we adopt Variational Score Distillation (VSD)~\cite{wang2024prolificdreamer}, which introduces a trainable LoRA module that incorporates camera information into the guidance. Its gradient can be written as:
\begin{small}
\begin{equation}
    \nabla L_{\mathrm{VSD}} = \mathbb{E}\left[\omega(t)\left(\hat{\epsilon}_{\phi}(z_t; y, t) - \hat{\epsilon}_{\mathrm{LoRA}}(z_t; y, t, c)\right) \frac{\partial x}{\partial \theta} \right],
\end{equation}
\end{small}
\noindent
where \(\hat{\epsilon}_{\mathrm{LoRA}}\) is the noise prediction of the LoRA-augmented branch and \(c\) denotes camera parameters. 
This case-specific LoRA is optimized jointly with the 3D object, leading to sharper details and more accurate geometry. 
In our pipeline, we adopt GaussianDreamer~\cite{yi2023gaussiandreamer} with VSD to obtain the base object \(O_1\), which provides geometry and coarse appearance for the subsequent stylization stage.

Although our main results use this VSD-based optimization pipeline, Stage~2 is agnostic to the specific 3DGS generator. 
In particular, end-to-end generators such as TRELLIS~\cite{trellis2024} can also be employed to produce \(O_1\); corresponding results are provided in the supplementary material.

\subsection{Stage 2: Style Distillation from Image}
\label{sec:stage2}

In the second stage, we transfer the style of a reference image \(I_s\) onto the base object \(O_1\) to obtain a stylized result \(O_s\) that is consistent with both the text prompt and the style image. 
We leverage a pre-trained 2D diffusion control model and an Image Condition Processor to extract style features and inject them into the diffusion guidance, forming our Variational Stylized Score Distillation (VSSD) framework (Figure~\ref{fig:sic3d}). 
At each optimization step, random camera views are rendered from \(O_1\), and VSSD guidance is computed using the text \(y\), style image \(I_s\), and camera parameters \(c\). 
Accumulating this guidance over multiple iterations produces the final stylized object \(O_s\). 
The following subsections introduce the IP-Adapter, VSSD formulation, and the scaling constraint used in Stage~2.

\subsubsection{IP-Adapter as Image Condition Processor}

To incorporate image-conditioned style information, we use a pre-trained IP-Adapter~\cite{ye2023ip} as the Image Condition Processor. 
It employs a decoupled cross-attention mechanism to jointly process text and image inputs. 
The style image \(I_s\) is first encoded by a CLIP image encoder into an embedding \(\mathbf{e}_I\), which is further projected by IP-Adapter into \(\mathbf{e}'_I\). 
Thus, the diffusion model is conditioned on both the text embedding \(\mathbf{e}_y\) and the processed style embedding \(\mathbf{e}'_I\).

In the \(i\)-th cross-attention layer, the queries \(Q^i\) are obtained from the previous feature map, while keys and values \((K_y^i, V_y^i)\) and \((K_I^i, V_I^i)\) are derived from text and image embeddings, respectively. 
The decoupled cross-attention output is defined as:
\begin{small}
\begin{equation}
    \boldsymbol{Z}_{\mathrm{mix}}^i = \operatorname{CA}(Q^i, K_y^i, V_y^i) + \lambda \cdot \operatorname{CA}(Q^i, K_I^i, V_I^i),
\end{equation}
\end{small}
\noindent
where \(\lambda\) is a hyper-parameter (IP-Adapter scale) controlling the relative contribution of image and text. 
Applying IP-Adapter across all cross-attention layers enables multi-modal conditioning, so that the final result reflects both semantic and stylistic cues.

\subsubsection{Variational Stylized Score Distillation (VSSD)}

Stage~2 focuses on enriching the surface texture of \(O_1\) by injecting style information in a viewpoint-consistent manner. 
To account for camera effects, we introduce a camera projection layer that maps camera parameters into embeddings \(\mathbf{e}_c\). 
With IP-Adapter providing style condition \(s\) and LoRA capturing object-specific information, we define the VSSD loss as:
\begin{small}
\begin{equation}
    \min_{o \in O} L_{\mathrm{VSSD}}(o) := \mathbb{E}\left[\omega(t)\big(\hat{\epsilon}_{\phi}(z_t; y, t, s) - \hat{\epsilon}_{\psi}(z_t; y, t, c)\big)\right],
\end{equation}
\end{small}
\noindent
where \(\hat{\epsilon}_{\phi}(z_t; y, t, s)\) is the prediction of the IP-Adapter-augmented diffusion model and \(\hat{\epsilon}_{\psi}\) is the prediction of the LoRA branch parameterized by \(\psi\).

During Stage~2, the LoRA parameters and the camera projection layer are trained jointly using a diffusion objective:
\begin{small}
\begin{equation}
    L_{\psi} = \mathbb{E}\big\|\hat{\epsilon}_{\psi}(z_t; \varnothing, t, c) - \epsilon\big\|^2.
\end{equation}
\end{small}

\noindent The camera projection layer is initialized from Stage~1, while the LoRA parameters are reinitialized to avoid bias from the coarse style in \(O_1\). 
By encoding object geometry into LoRA, VSSD produces geometry-consistent and style-aligned guidance across views. 
The degree of stylization is controlled by the IP-Adapter scale and the timestep schedule; ablations are provided in the supplementary material.

\subsubsection{Scaling Constraint for 3DGS}
\label{sec:vc}

\begin{figure}[tbh]
    \centering
    \includegraphics[width=0.75\linewidth]{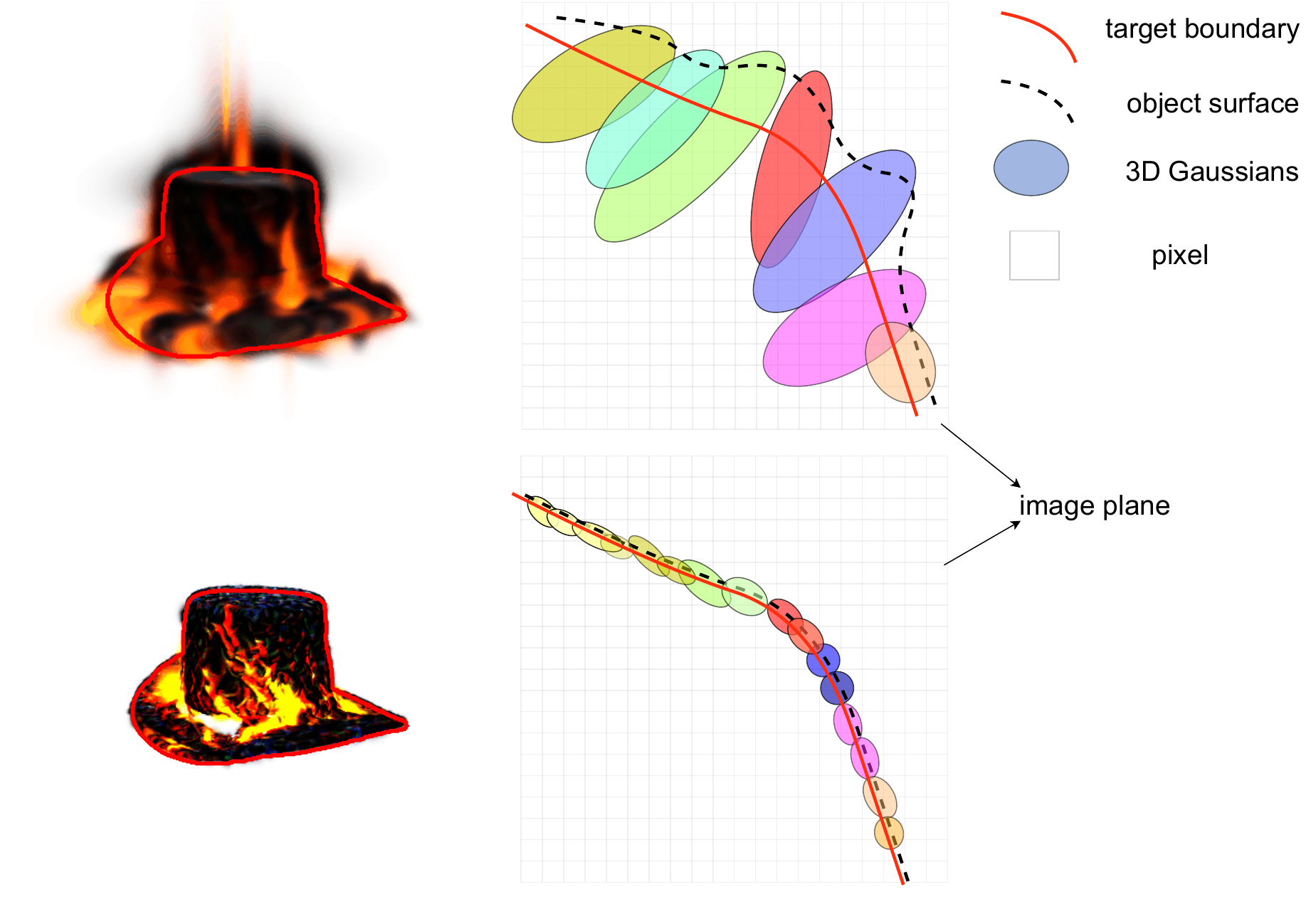}
    \caption{Impact of different scales of Gaussians. The red solid line indicates the target boundary, while the black dashed line represents the object surface. In the top row, where the Gaussian scale is too large, significant artifacts appear, even when the Gaussian centers are near the target boundary, leading to blurry boundaries. In contrast, the bottom row shows that smaller scales lead to higher fidelity and smoother contours.
    }
    \label{img:vc}
\end{figure}

During optimization, individual Gaussians may expand to cover larger regions on the image plane, which can cause artifacts and blurry stylization, as illustrated in Figure~\ref{img:vc}. 
To prevent this, we regularize the scale of each primitive so that its projected influence remains bounded and consistent.

Directly constraining the 3D volume of each Gaussian is insufficient, as highly anisotropic primitives can still exert large influence on the image plane despite having small volumes.
Alternatively, explicitly counting the number of affected pixels leads to non-differentiable objectives, making optimization impractical.
To address these limitations, we regularize the approximated surface area of each Gaussian, which provides a continuous proxy that better correlates with its visible contribution in rendered views.

We approximate the surface area of each anisotropic Gaussian by the ellipsoid formula
\begin{equation}
    A \approx 4 \pi\left(\frac{a^p b^p + a^p c^p + b^p c^p}{3}\right)^{\frac{1}{p}}, \quad p = 1.6075,
\end{equation}
where \(a, b, c\) are the scaling parameters along the principal axes and \(p\) is empirically chosen~\cite{Michon:NumericanaEllipsoidSurfaceArea}. 
We then penalize the variance of the logarithmic surface area:
\begin{equation}
    \mathcal{L}_{\mathrm{surface}} = \frac{1}{N} \sum_{i=1}^N\left(\log A_i - \log \bar{A}\right)^2,
\end{equation}
where \(A_i\) is the area of the \(i\)-th Gaussian and \(\bar{A}\) is the mean across all Gaussians. 
This scale-invariant regularization encourages a uniform distribution of influence among primitives.

Minimizing \(\mathcal{L}_{\mathrm{surface}}\) under fixed viewpoints leads to stable optimization, reduces view-dependent artifacts, and yields stylized outputs with sharper contours and clearer textures.

\subsection{Discussion}
\label{sec:discussion}

We briefly discuss several design choices in SIC3D. 
First, although IP-Adapter is originally used in a forward generative setting starting from noise, we repurpose it for gradient-based 3D optimization by feeding perturbed 3D renderings. 
Empirically, gradients on background regions have negligible impact, while object regions receive stable and style-consistent guidance, making the adapter effective for 3D stylization.

Second, directly applying 2D style transfer methods (e.g., AdaIN~\cite{huang2017arbitrary}) on rendered views often conflicts with the content induced by the text prompt, leading to incoherent textures. 
In contrast, diffusion-based guidance with IP-Adapter integrates style and text conditions within a single generative prior, producing more semantically faithful stylized 3D results.

Finally, the LoRA-adapted backbone, together with full-range timestep and camera sampling, helps balance stylization strength and detail preservation. 
Overall, these components make IP-Adapter a robust conditional guidance module for 3D stylization beyond its original generative setting.

\section{Experiments}

We compare SIC3D with a stylized-prompt baseline and three recent 3D style transfer methods: StyleGaussian~\cite{liu2024stylegaussian}, G-Style~\cite{kovacs2024}, and SGSST~\cite{galerne2024sgsst}. We report both qualitative and quantitative results, and ablation studies on key design choices.

\begin{figure*}[tbh]
  \centering
  \includegraphics[width=0.9\textwidth]{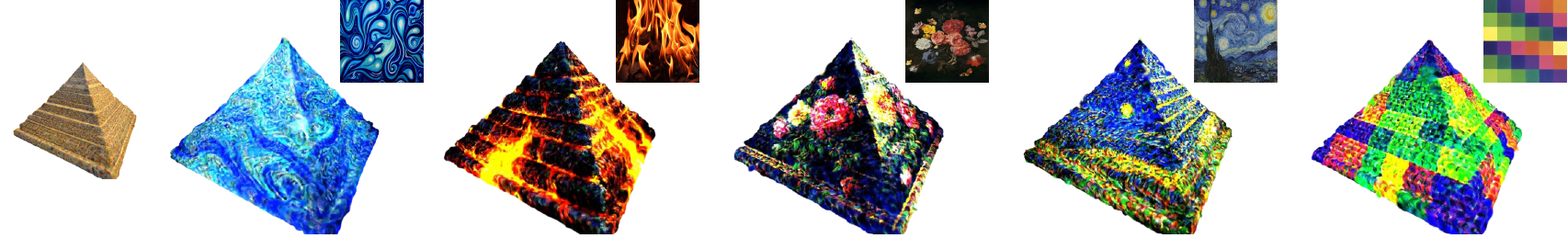}
  \caption{
  Results from SIC3D. 
  The object in the left column is generated with the prompt 
  ``An ancient Egyptian pyramid'' 
  The corresponding style images are shown in the top right of each sample. More visual results can be found in the Supplementary Material.
  }
  \label{fig:results}
\end{figure*}

\subsection{Implementation Details}

We follow GaussianDreamer~\cite{yi2023gaussiandreamer} and ProlificDreamer~\cite{wang2024prolificdreamer} to obtain the base object in the \emph{Object Generation Stage}, implemented within the ThreeStudio framework~\cite{threestudio2023} for both stages. 
All experiments are conducted with Stable Diffusion v1.5 as the underlying 2D prior. 
In Stage~2, we use the pre-trained IP-Adapter model \texttt{h94/IP-Adapter}, which is compatible with Stable Diffusion v1.5. 
We render four fixed viewpoints distributed evenly around the object with identical distance and elevation.

All experiments are performed on a single NVIDIA A100 GPU. 
For Stage~1, we primarily use GaussianDreamer with VSD for 500 optimization steps to obtain a robust base object \(O_1\). 
As an alternative, TRELLIS~\cite{trellis2024} can generate \(O_1\) in about 20 seconds at the cost of reduced robustness; corresponding results are provided in the supplementary material (Section~\ref{sec:more_results}). The second stage runs for 500 steps in our main experiments; in practice, visually plausible stylization is often achieved after around 200 steps. Runtime and memory consumption detail can be found in the Supplementary Material (Section~\ref{sec:resource}).

\subsection{Comparison Methods \& Evaluation Metrics}

\paragraph{Stylized-prompt baseline}
As a strong baseline for text+image conditioning, we construct a \emph{stylized prompt} by appending a style description extracted from the style image to the original text. 
We use GPT-4o to caption each style image, focusing on salient stylistic attributes (details in the supplementary material). 
The resulting description \(\hat{s}\) is appended to the text prompt \(y\) to form \(y' = y + \text{``in the style of } \hat{s}\text{''}\), which is then used in a single-stage 3DGS generation with VSD. 
For example, the prompt ``a fox'' with a fire image becomes ``a fox in the style of dancing flames, with bright orange and yellow fire dancing and twisting'' for generation.

\paragraph{3D style transfer baselines}
We compare SIC3D to three recent 3DGS style transfer methods. 
StyleGaussian~\cite{liu2024stylegaussian} adapts Gaussian colors to a 2D style image using VGG-based features and a 3D CNN decoder, optimizing only color parameters. 
G-Style~\cite{kovacs2024} stylizes entire 3DGS scenes by minimizing CLIP-S and NNFM losses with multi-view constraints and color matching. 
SGSST~\cite{galerne2024sgsst} uses a multiscale SOS loss to capture global neural statistics while selectively updating Gaussian colors.

\paragraph{Evaluation metrics}
We adopt NNFM, RMSE, and LPIPS following prior work. 
To avoid bias from empty backgrounds, all metrics are computed on masked object regions only. 
Each object is rendered from four fixed viewpoints and the metrics are averaged. 
NNFM measures style similarity via Gram matrices of VGG19 feature maps, with lower scores indicating closer adherence to the style image. 
RMSE and LPIPS are computed between  renderings and the style image: RMSE captures pixel-level deviations, while LPIPS reflects perceptual differences in feature space.

\begin{figure*}[tbh]
    \centering
    \hspace{1cm}
    \includegraphics[width=\linewidth]{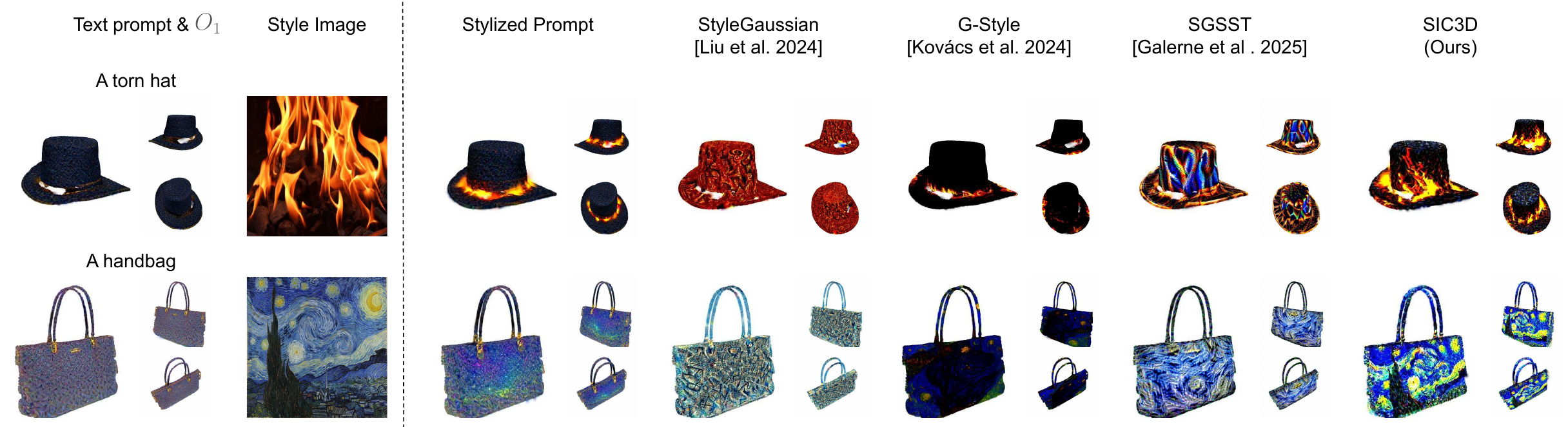}
    \caption{
    Qualitative comparison between SIC3D, the stylized-prompt baseline, and state-of-the-art 3D style transfer methods (G-Style~\cite{kovacs2024}, StyleGaussian~\cite{liu2024stylegaussian}, SGSST~\cite{galerne2024sgsst}). 
    Each example shows three renderings from different viewpoints. More examples are given in the Supplementary Material.
    }
    \label{fig:comparison}
\end{figure*}

\subsection{Qualitative Analysis}

Figure~\ref{fig:comparison} compares SIC3D with the baseline and SOTA methods under various text–style image pairs; additional examples are provided in the supplementary material. 
SIC3D consistently transfers characteristic style patterns from the reference image while preserving the underlying geometry. 
Compared with the stylized-prompt baseline, which often distorts geometry due to ambiguous textual descriptions, our two-stage strategy separates geometry from appearance and yields stronger stylization with better structural integrity. 
Relative to StyleGaussian and G-Style, SIC3D captures finer details from the style image and maintains smoother, more coherent textures across the surface. 
While SGSST enforces strong style consistency, it frequently produces artifacts that disrupt texture and geometry, whereas SIC3D better balances stylization strength and geometric plausibility.

\subsection{Quantitative Analysis}

For quantitative evaluation, we compute NNFM, RMSE, and LPIPS on masked renderings from four fixed viewpoints. 
NNFM is based on Gram matrices of VGG19 feature maps (`conv1\_1', `conv2\_1', `conv3\_1', `conv4\_1', `conv5\_1') and measures the style discrepancy between each rendering and the style image; lower values indicate stronger style adherence. 
RMSE and LPIPS are computed with respect to the style image, reflecting pixel-level deviations and perceptual differences, respectively.

As summarized in Table~\ref{tab:clip-s}, SIC3D achieves the lowest NNFM score (0.151), indicating the best style alignment among all methods. 
It also obtains competitive RMSE (1.301), demonstrating that stylization is applied in a controlled manner, and the best LPIPS score (0.877), suggesting strong perceptual plausibility. 
Overall, SIC3D provides the most effective and balanced style transfer.

\begin{table}[tbh]
    \caption{
    Quantitative comparison on masked renderings in terms of NNFM, RMSE, and LPIPS. 
    Bold indicates the best performance for each metric.
    }
    \begin{center}
      \footnotesize
        \label{tab:clip-s}
        \begin{tabular}{@{}lccc@{}}
            \hline
            Method & NNFM $\downarrow$ & RMSE $\downarrow$ & LPIPS $\downarrow$ \\ \hline
            Stylized Prompt & 0.156   & 1.413 & 0.881 \\ 
            StyleGaussian   & 0.159   & \textbf{1.232} & 0.882 \\ 
            G-Style         & 0.154   & 1.390 & 0.881 \\ 
            SGSST           & 0.153   & 1.417 & 0.889 \\ 
            SIC3D           & \textbf{0.151}  & 1.301 & \textbf{0.877} \\ \hline
        \end{tabular}
    \end{center}
\end{table}

Furthermore, we use GPT-4o as a proxy for human evaluation, following~\cite{wu2024gpt}. 
We design comparison instructions focusing on \emph{Object Quality} (five aspects as in~\cite{wu2024gpt}) and \emph{Style Alignment} with the reference image; details are provided in the supplementary material. 
We randomly sample 100 combinations from 20 text prompts and 9 style images (4 realistic, 5 artistic), and generate 3 random seeds per combination. 
Each method is compared pairwise 350 times by GPT-4o across all text–image–seed pairs, and we compute ELO scores~\cite{elo1967proposed}.

\begin{table}[tbh]
  \caption{
  GPT-4o evaluation (ELO scores) for Object Quality and Style Alignment across different stylized 3DGS generation methods. 
  Higher is better; bold indicates the best score.
  }
  \begin{center}
      \footnotesize
      \begin{tabular}{@{}lcc@{}}
        \toprule
        Method & Object Quality & Style Alignment \\
        \midrule
        Stylized Prompt & 1044.600 & 878.905   \\
        StyleGaussian~\cite{liu2024stylegaussian} & 925.368  & 848.153   \\
        G-Style~\cite{kovacs2024} & 1015.814 & 893.972   \\
        SGSST~\cite{galerne2024sgsst} & 983.604  & 980.170   \\
        SIC3D & \textbf{1075.125} & \textbf{1154.929} \\
        \bottomrule
      \end{tabular}
      \label{tab:comparison}
  \end{center}
\end{table}

SIC3D achieves the highest ELO scores in both \emph{Object Quality} and \emph{Style Alignment}, indicating that it provides the best trade-off between 3D plausibility and faithful style transfer.

\subsection{Ablation Study}
\label{sec:ablation}

We ablate three key components of SIC3D: the scaling constraint, fixed camera sampling, and LoRA in Stage~2.

\begin{figure}[htb]
    \centering
    \includegraphics[width=0.75\linewidth]{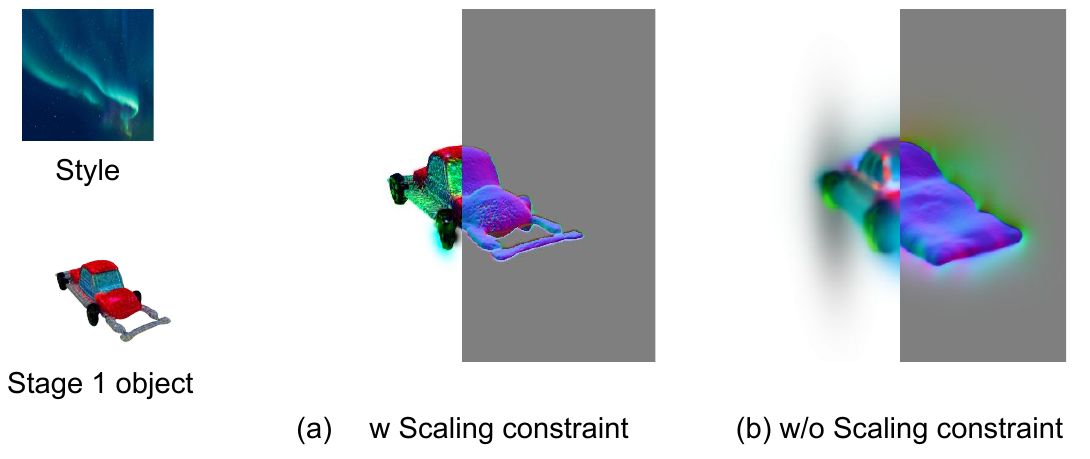}
    \caption{
    Ablation on the scaling constraint. 
    (a) With the constraint, Gaussians remain compact, leading to sharp boundaries and stable geometry. 
    (b) Without it, Gaussians expand, causing artifacts and blurred contours.
    In each case, we show RGB rendering on the left and normal map on the right.
    }
    \label{fig:ablation}
\end{figure}

\paragraph{Scaling constraint}
To assess the impact of the scaling constraint, we vary the average Gaussian scale during optimization. 
As shown in Figure~\ref{fig:ablation}, enabling the constraint keeps Gaussian radius small (3–4 pixels), yielding clean boundaries and plausible geometry. 
Without it, Gaussians grow excessively (40–60 pixels), aggregating gradients over large regions and producing distorted geometry and artifacts.

\begin{figure}[tbh]
    \centering
    \includegraphics[width=0.75\linewidth]{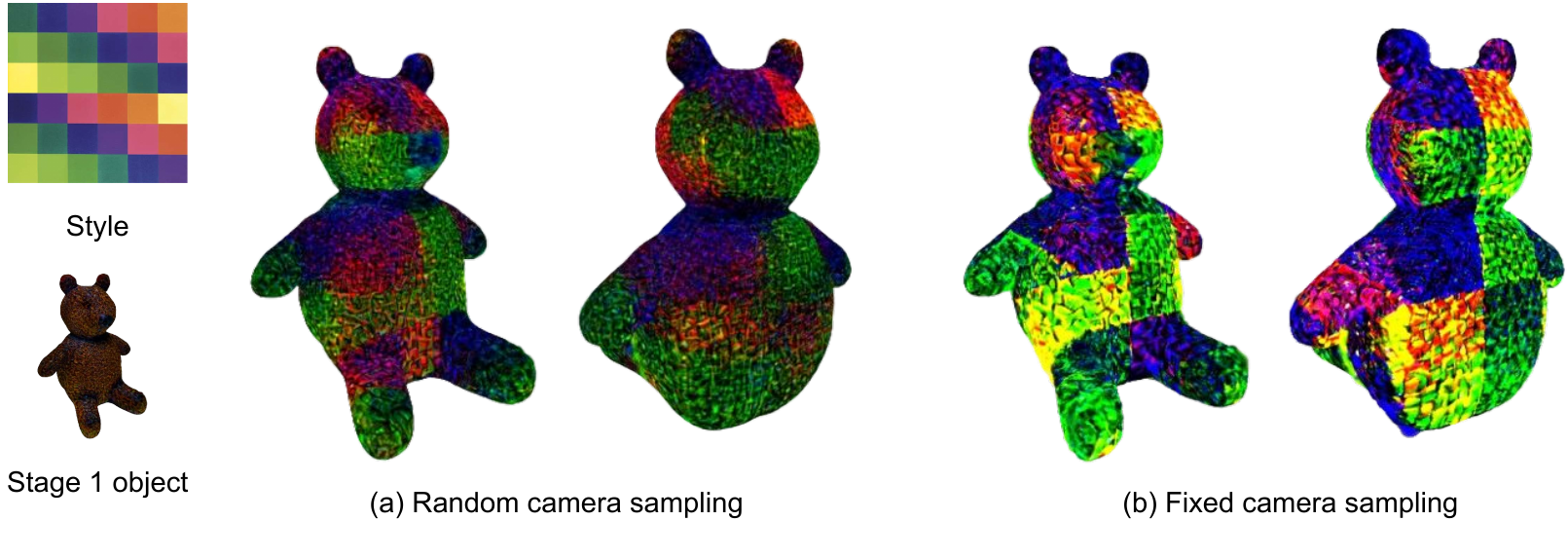}
    \caption{
    Ablation on camera sampling. 
    (a) Random viewpoints yield lower object quality and weaker surface style patterns. 
    (b) Fixed viewpoints result in clearer geometry and more stable texture stylization.
    }
    \label{fig:ablation_camera}
\end{figure}

\paragraph{Camera sampling}
We compare random and fixed camera sampling strategies in Stage~2. 
As shown in Figure~\ref{fig:ablation_camera}, random viewpoints lead to weaker and less coherent surface patterns, especially for dense or repetitive styles, whereas fixed viewpoints produce clearer geometry and more consistent stylization across views.

\begin{figure}[tbh]
    \centering
    \includegraphics[width=0.75\linewidth]{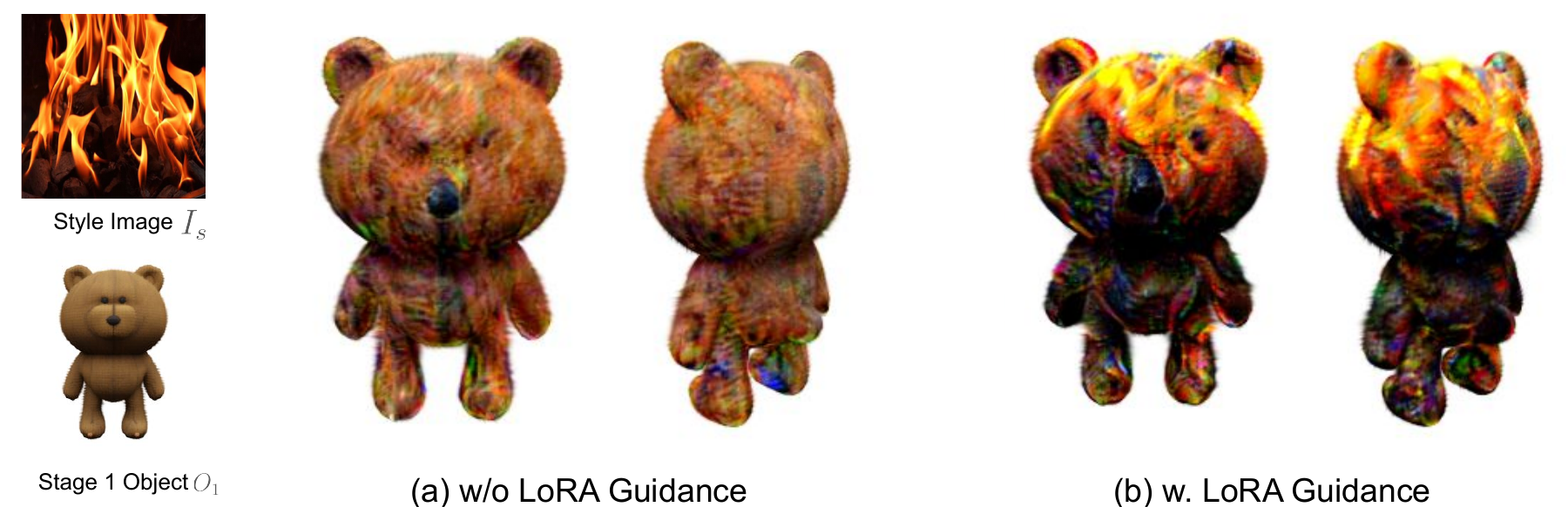}
    \caption{
    Ablation on LoRA. 
    (a) Without LoRA, surface style pattern are insignificant. 
    (b) With LoRA, surfaces reflect strong style patterns from \(I_s\).
    }
    \label{fig:ablation_lora}
\end{figure}

\paragraph{LoRA layers}
Figure~\ref{fig:ablation_lora} shows the effect of LoRA in Stage~2. 
Without LoRA, the surfaces appear rough and the transferred patterns are weak. 
With LoRA, the results exhibit clear patterns resembling the style image, demonstrating that LoRA is crucial for faithful and consistent stylization.

\section{Conclusions}

In this paper, we introduce a novel pipeline for two-stage stylized 3DGS object generation, SIC3D, which is guided by both image and text conditions. We propose using VSSD loss to produce stylized 3D results. To address artifacts that may arise in the generation process, we identify its root cause within the guidance provided by the diffusion model and propose a background gradient control strategy, effectively suppressing the formation of additional meaningless 3D structures. Comparative experiments under standard metrics such as NNFM, RMSE and LPIPS show a superior result to compared methods, while complementary GPT-4o evaluations further validate SIC3D's effectiveness. In addition to object generation, SIC3D naturally generalizes to broader applications including stylized 3D scene synthesis and 3D style transfer, highlighting its flexibility and potential for future research.

\textit{Limitations and future work: }
Despite the improvements introduced by our fixed-viewpoint strategy, the method still relies on a pretrained text-to-image diffusion model lacking explicit multi-view consistency. This can lead to localized texture artifacts in overlapping regions between fixed views. Future work will explore the integration of multi-view consistent diffusion models \cite{shi2023mvdream,  long2024wonder3d, shi2023zero123++} to enhance cross-view style consistency and further improve the generation quality.

\bibliographystyle{IEEEbib}
\bibliography{main}

\clearpage
\centerline{\noindent {\centering\huge \textbf{Appendix}}}
\vspace{10pt}
\setcounter{page}{1}

\setcounter{section}{0}

In this supplementary material, we provide additional results and detailed explanations of our methods. Section \ref{sec:structure} gives an overview of the structure of our network. Section \ref{sec:complexity} provides an analysis of the complexity of style image descriptions of the style image and its effect on the baseline. Section \ref{sec:more_ablation} shows more results of the ablation study in Section \ref{sec:ablation}. Section \ref{sec:instantstyle} presents additional comparisons with InstantStyle and backbone generalization results. Finally, Section \ref{sec:more_results} gives more results of SIC3D generation.

\subsection{Network Structure}
\label{sec:structure}
Our generation model can be divided to several parts: diffusion model, IP-Adapter, LoRA and camera projection layer. The design of the diffusion model follows the pre-trained model, which is \textit{Stable Diffusion v1.5}. IP-Adapter is also pre-trained, compatible with the chosen diffusion model, and specifically trained for the stylization task. It is plugged into each of the cross-attention layers of the denoising U-Net \cite{ye2023ip}. Different designs of LoRA and the camera projection layer have an insignificant effect on the final performance, therefore we use the same design as ProlificDreamer \cite{wang2024prolificdreamer}.

\subsection{Runtime and Resource Requirements}
\label{sec:resource}
We report the computational resources required by SIC3D during both stages of optimization, and compare the full model with LoRA against a variant without LoRA.

\paragraph{GPU memory usage}
Table~\ref{tab:mem} summarizes the peak GPU memory usage for Stage~1 initialization and Stage~2 stylization. The LoRA-enabled variant introduces additional parameters in the cross-attention layers, resulting in slightly higher GPU consumption, particularly during Stage~2 where gradients must be maintained for LoRA updates. Despite this overhead, both variants remain well within the capacity of a single 80GB GPU.

\begin{table}[htb]
  \centering
  \footnotesize
  \caption{Estimated peak GPU memory usage (in GB) with and without LoRA.}
  \label{tab:mem}
  \begin{tabular}{lcc}
    \toprule
    \textbf{Setting} & \textbf{Stage 1} & \textbf{Stage 2} \\
    \midrule
    Ours (with LoRA)    & 18.7 & 18.9 \\
    Ours (without LoRA) & 9.58 & 9.8 \\
    \bottomrule
  \end{tabular}
\end{table}

\paragraph{Per-step runtime}
We measure the per-iteration runtime for both stages, summarized in Table~\ref{tab:time}. Stage~1 is dominated by view-conditioned VSD optimization and 3D Gaussian rendering. Stage~2 additionally invokes LoRA-weighted U-Net forward and backward passes, leading to a moderate increase in computation time. Overall, LoRA adds only a small runtime overhead and does not affect the practicality of the system.

\begin{table}[htb]
  \centering
  \footnotesize
  \caption{Estimated per-step runtime (in seconds) with and without LoRA.}
  \label{tab:time}
  \begin{tabular}{lcc}
    \toprule
    \textbf{Setting} & \textbf{Stage 1} & \textbf{Stage 2} \\
    \midrule
      Ours (with LoRA)    & 0.66 & 0.85 \\
    Ours (without LoRA) & 0.38 & 0.59 \\
    \bottomrule
  \end{tabular}
\end{table}

Overall, the full method with LoRA introduces slight computational overhead in both memory usage and runtime, but the benefits of parameter-efficient finetuning outweigh the added cost. The method remains efficient and easily runnable on a single high-memory GPU.

\subsection{Stylized Prompt Complexity Analysis}
\label{sec:complexity}
\begin{figure*}[tb]
    \centering
    \includegraphics[width=0.85\linewidth]{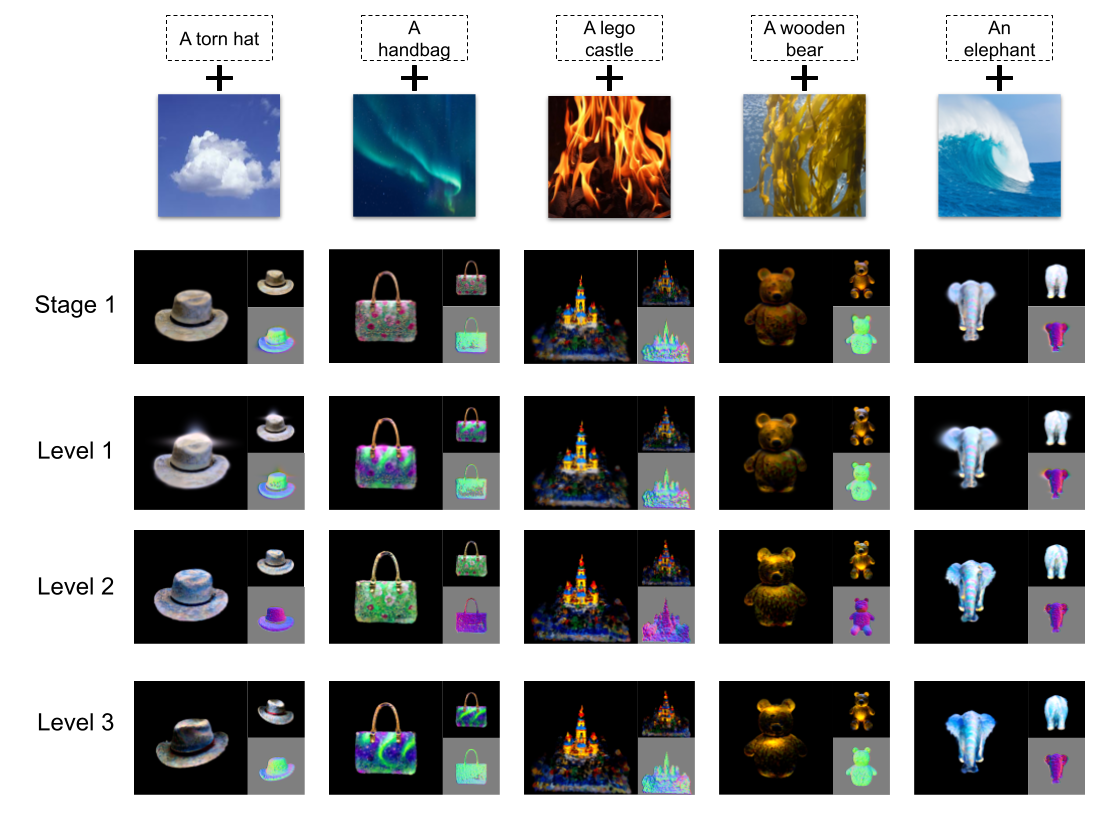}
    \caption{Generation results for baseline with different complexity level of stylized prompt (see Table \ref{tab:sp})}
    \label{fig:complexity_2}
\end{figure*}

In this section, we provide an analysis regarding the baseline, focusing on the complexity level of the stylized prompt. Each style image will be described by the GPT-4o model, focusing on the stylistic elements of the image. However, descriptions with different levels of detail reflect different levels of style information as shown in Figure \ref{fig:comparison}. In Table \ref{tab:sp}, we present different levels of complexity of style caption. In Figure \ref{fig:complexity_2}, we perform a comparison of different complexity style prompts for all the baseline examples that appeared in Figure \ref{fig:comparison}. We can see that Level 3 complexity of the style prompt provides the most information from the style image and has the best style alignment with the style image. Therefore, we use complexity level 3 for our comparisons in Figure \ref{fig:comparison} in the main paper.

\begin{table*}[tb]
  \centering
  \footnotesize
  \caption{Different level of complexity for style image description.}
  \label{tab:sp}
  \begin{tabularx}{\textwidth}{@{}>{\raggedright\arraybackslash}X>{\raggedright\arraybackslash}X>{\raggedright\arraybackslash}X>{\raggedright\arraybackslash}X>{\raggedright\arraybackslash}X>{\raggedright\arraybackslash}X@{}}
    \toprule
    Complexity Level & cloud & northern light & fire & seaweed & wave \\
    \midrule
    Level 1 & fluffy cloud & aurora borealis & dancing flames & sunlit seaweed & massive waterwave \\
    \midrule
    Level 2 & fluffy, soft clouds against a deep blue sky. & graceful aurora with soft, ethereal green hues & vivid, dynamic flames with smooth, flowing shapes & golden, wavy seaweed with soft underwater lighting & majestic wave with crisp, dynamic curling motion \\
    \midrule
    Level 3 & fluffy white cloud floating peacefully against a deep blue sky, evoking a sense of calm and openness. & mesmerizing aurora borealis with sweeping green and blue waves, dancing across a star-studded night sky. & dancing flames, with bright orange and yellow fire dancing and twisting & long, golden-brown seaweed strands floating underwater, illuminated by sunlight filtering through the water. & powerful ocean wave cresting with intense energy, cascading white foam over vivid blue water under a clear sky. \\
    \bottomrule
  \end{tabularx}
\end{table*}


\subsection{GPT-4o Evaluation Detail}
We designed instructions for GPT-4o to evaluate style alignment between object rendering and style image. The details for each instruction is listed below:

\begin{small}
\lstset{
    breaklines=true, 
    xleftmargin=0pt, 
    framexleftmargin=0pt,
    showstringspaces=false
}
\begin{lstlisting}

Our task is to compare two 3D models to see if they match the style of a reference image. Both the 3D models are generated from the same base text description and influenced by the style feature of the reference image. We will evaluate the models based on the aspects of style alignment with the reference image.

I will provide you with a multi-part image divided into three rows and four columns, each of which contains an image. In the first two rows, the four images on the left are four renderings of object 1 from different viewpoints, and the four images on the right are four renderings of object 2 from different viewpoints. The last row has four identical images of the style reference being used.

# Instruction
Examine the multi-part image carefully. Attempt to mentally construct a 3D model of object 1 using the 4-view rendering in the left part. Attempt to mentally construct a 3D model of object 2 using the 4-view rendering in the right part. For each 3D model, determine which one has a similar style to the reference image. An ideal model's rendering should have strong commonality with reference image, especially in the colour palette and continuity of tones. For example, the reference image contains fire with intense orange-red tones. Desired object's rendering should also have a similar color and tone as if it's also under fire. While there are subtle differences in flame shape and detailed textures, the overall visual effect is very consistent.
Take a really close look at each of the renderings for the two 3D objects and determine which one is better aligned to the reference image.

# Output format
Provide a short analysis before giving a final answer. The analysis should be very concise and accurate.
Your answer should be one of the  following three options:
1. Left (object 1) is better;
2. Right (object 2) is better;
3. Cannot decide. 
IMPORTANT: PLEASE USE THE THIRD OPTION SPARSELY.
In the last row, give your final decision in a single number.

An example output looks like the following: 

Analysis:
Style Alignment: The left one xxxx; The right one xxxx; The left/right one is better or I cannot decide.
Final answer: 
x (e.g., 1 / 2 / 3) 
\end{lstlisting}
\end{small}

\begin{figure}[tb]
    \centering
    \includegraphics[width=\linewidth]{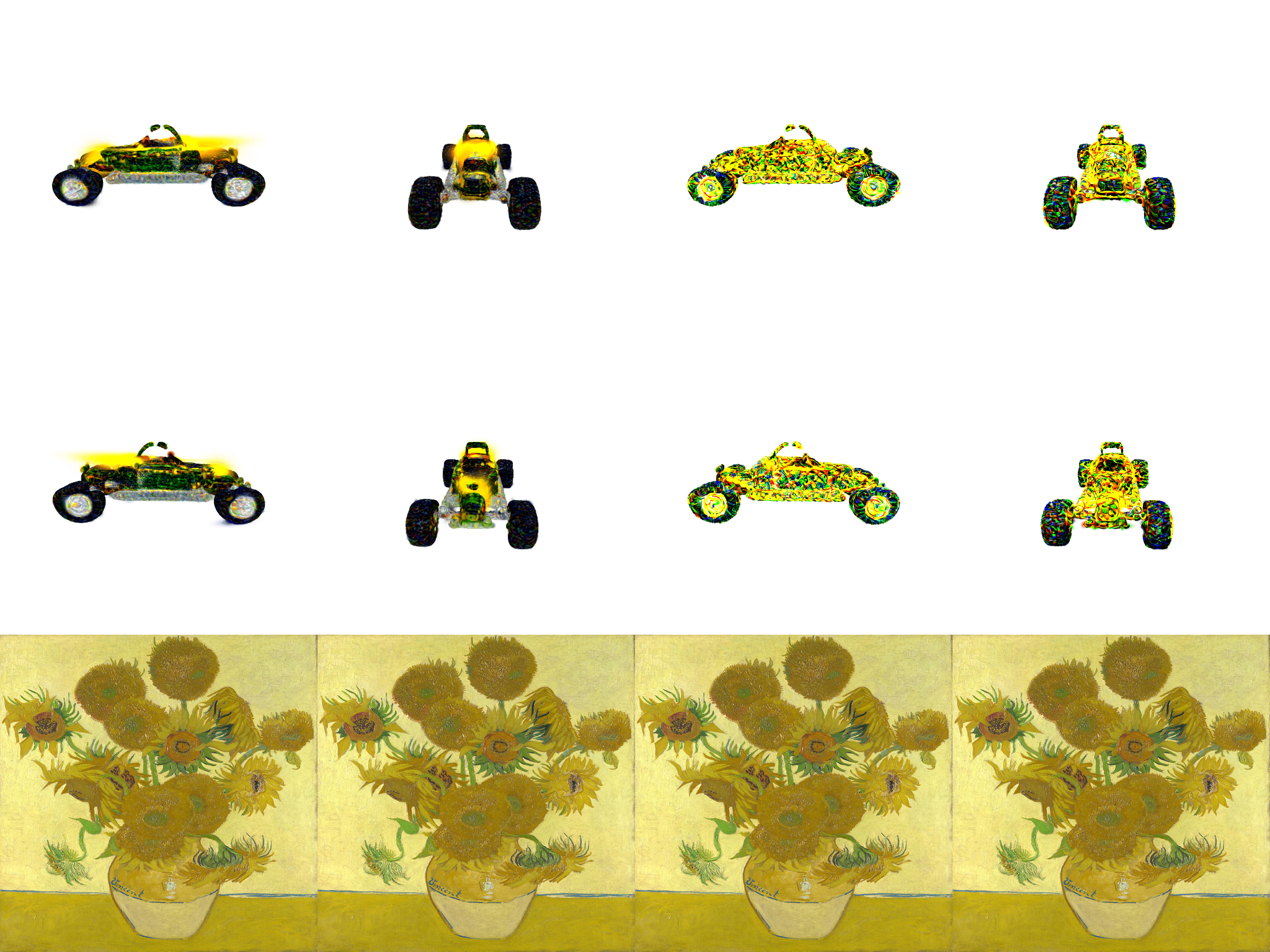}
    \caption{Input for Style Alignment Evaluation. Left part result is produced from baseline style prompt, the right part is produced from SIC3D pipeline.}
    \label{fig:input}
\end{figure}

\noindent An example answer for Figure \ref{fig:input} from GPT-4o is as follows:

\begin{small}
\lstset{breaklines=true}
\begin{lstlisting}
Analysis:
Style Alignment: The left object (1) has softer colors with prominent yellows and greens, somewhat resembling the sunflower tones. The right object (2) exhibits more intricate and varied colors, closely mimicking the brushstroke texture and color variation of the reference image. The right one is better aligned with the reference image's style.

Final answer:
2
\end{lstlisting}
\end{small}

\subsection{More Ablation Study}
\label{sec:more_ablation}
In this section, we present more results for the Ablation Study in Section \ref{sec:ablation} in the main paper. The aspects we focus on are the effect of Stage 2 Steps, IP-Adapter scale and timestep sampling range. 






\subsection{Stage 2 Steps}
\begin{figure*}[tb]
    \centering
    \includegraphics[width=1\linewidth]{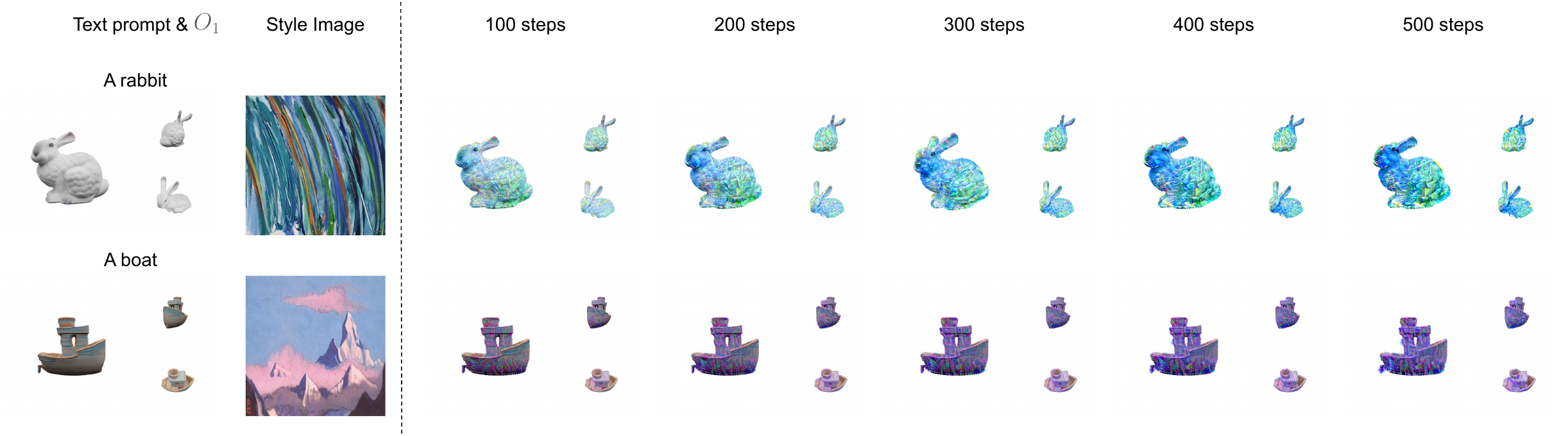}
    \caption{Multi-view generation results of two examples (“A rabbit”, “A boat” and two different style images) with different numbers of optimization steps in Stage~2 (100, 200, 300, 400, and 500). Each row corresponds to one text prompt–style image pair, and each column shows the result at a specific step.}
    \label{fig:opt_steps}
\end{figure*}

To investigate the effect of optimization steps of Stage~2, we present multi-view renderings of the generated objects at different optimization steps, ranging from 100 to 500. As shown in Figure~\ref{fig:opt_steps}, the overall texture pattern is already determined at the early stage, while the degree of stylization becomes more stable after around 200 steps. Beyond this point, the optimization primarily refines local texture patterns and improves consistency across views. For clarity and stability of stylization, all the results presented in the main paper are obtained with 500 optimization steps.

\subsection{IP-Adapter scale}
\begin{figure*}[tb]
    \centering
    \includegraphics[width=1\linewidth]{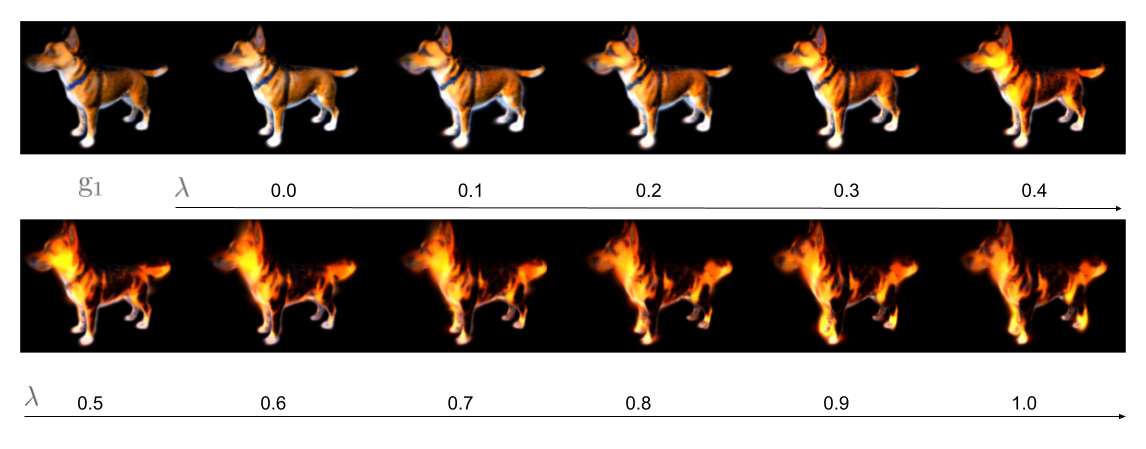}
    \caption{Influence of IP-Adapter scale. First image is the rendering of first stage result \(\mathrm{g}_1\) of prompt "a dog". Then we show the result of applying different value of IP-Adapter scale (within range [0.0, 1.0]) with the style image of fire.}
    \label{fig:ip_adapter_scale}
\end{figure*}

We present the influence of different IP-Adapter scale in Figure \ref{fig:ip_adapter_scale}. From the image, we can see that when the IP-Adapter scale is smaller than 0.5, the influence from the style image is not obvious; when it is larger than 0.5, the style features from the style image become more and more apparent, gradually covering the original texture.

\begin{figure*}[tb]
    \centering
    \includegraphics[width=1\linewidth]{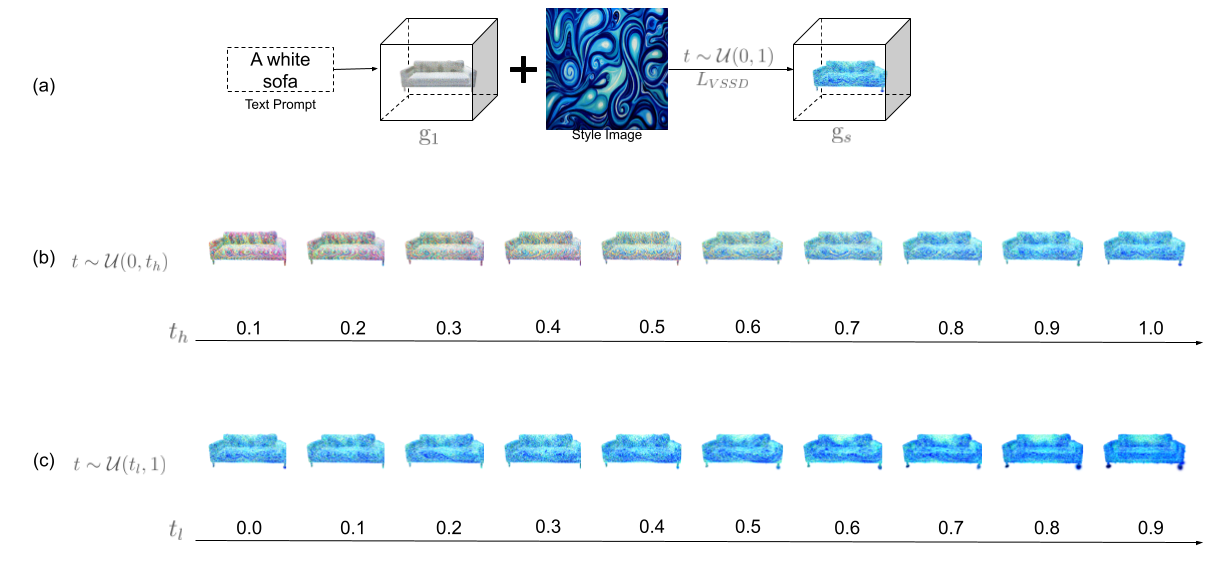}
    \caption{Influence of timestep sampling range. Row (a) represents the generation process with timestep sampled from 0 to 1. Row (b) represents the results for different the upper limit, from 0 to \(t_h\). \(t_h\) range from 0.1 to 1.0. Row (c) are the results for different lower limit, from \(t_l\) to 1.0. \(t_l\) range from 0.0 to 0.9.}
    \label{fig:noise_scale}
\end{figure*}

\subsection{Timestep sampling range}
In image style transfer task based on diffusion model, a random noise will be added to the original image for \(t\) timesteps, which represents the noise scale. Perturbed image will be denoised to obtain the stylized image. Noise scale added to the original image determines the degree of stylization. If a low scale of noise is added, most of the information from the original image is retained, resulting in a less significant effect on the image. If a high scale of noise is added to the image, most of the information from the original image is corrupted. The denoised image will contain more features from style image. 

In 3D stylization task, noise scale also plays a crucial role in stylization. By controlling the timestep sampling range, we can control and analyze the degree of stylization and the quality of object. Our method use a random sampling strategy between all the timesteps, \(t \sim \mathcal{U}(0, 1)\). We present a range of comparison of different timestep sampling range in Figure \ref{fig:noise_scale}. \(t_l\) and \(t_h\) represent the lower bound percentage and upper bound percentage of the total timestep range. (e.g. \(t_l\) is 0.1 and total timestep is 1000, then the lower bound for timestep range is 100.) We find that the guidance provided by the diffusion model focuses on changing the style of the object at high noise scale, while it focuses on optimizing the details at low noise scale.


\subsection{Comparison of Stage~1 Initialization Methods}

\begin{figure*}[tb]
    \centering
    \includegraphics[width=1\linewidth]{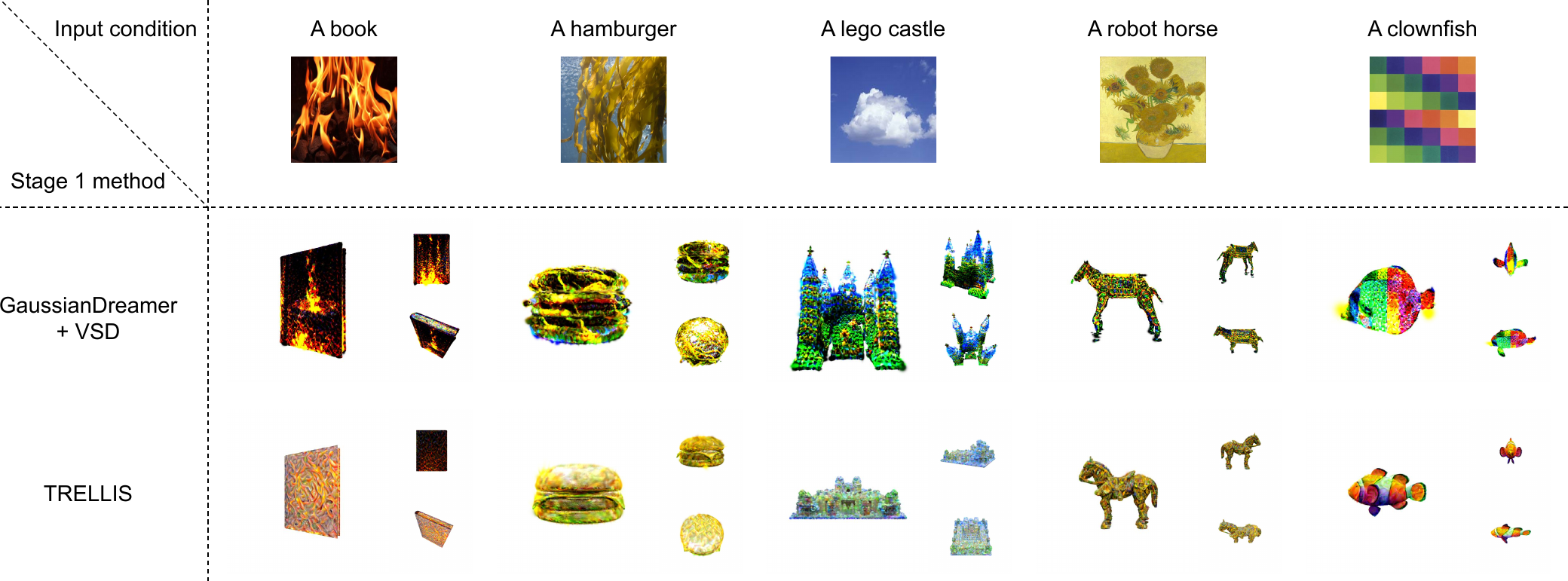}
    \caption{Comparison results on different generation model in Stage 1.}
    \label{fig:stage1_comparison}
\end{figure*}

Figure~\ref{fig:stage1_comparison} compares two representative Stage~1 generation methods, GaussianDreamer \cite{yi2023gaussiandreamer} with VSD loss \cite{wang2024prolificdreamer} and Trellis \cite{trellis2024}, given the same text prompts. Both approaches are able to produce objects that are consistent with the textual descriptions, as shown in the first-stage outputs. When proceeding to Stage~2, the degree of stylization and the way it fuses with the underlying geometry remain comparable between the two initialization strategies. This demonstrates that our Style Distillation Stage can adapt to different Stage~1 backbones without being sensitive to the specific initialization choice.

\subsection{Comparison with InstantStyle and Backbone Generalization}
\label{sec:instantstyle}

To further validate the effectiveness of our IP-Adapter conditioning strategy, we compare SIC3D against the layer selection strategy proposed in InstantStyle\cite{DBLP:journals/corr/abs-2404-02733}, which restricts style injection to specific cross-attention layers. As shown in Table~\ref{tab:instantstyle}, applying the InstantStyle layer selection strategy in our 3D optimization setting leads to weaker stylization across all three metrics, whereas our full-layer conditioning consistently achieves better style adherence.

\begin{table}[tb]
  \centering
  \footnotesize
  \caption{Quantitative comparison between InstantStyle layer selection strategy and our method on masked renderings.}
  \label{tab:instantstyle}
  \begin{tabular}{lccc}
    \toprule
    \textbf{Method} & \textbf{NNFM} $\downarrow$ & \textbf{RMSE} $\downarrow$ & \textbf{LPIPS} $\downarrow$ \\
    \midrule
    InstantStyle & 0.160 & 1.418 & 0.886 \\
    Ours         & \textbf{0.151} & \textbf{1.301} & \textbf{0.877} \\
    \bottomrule
  \end{tabular}
\end{table}

\begin{figure}[tb]
    \centering
    \includegraphics[width=\linewidth]{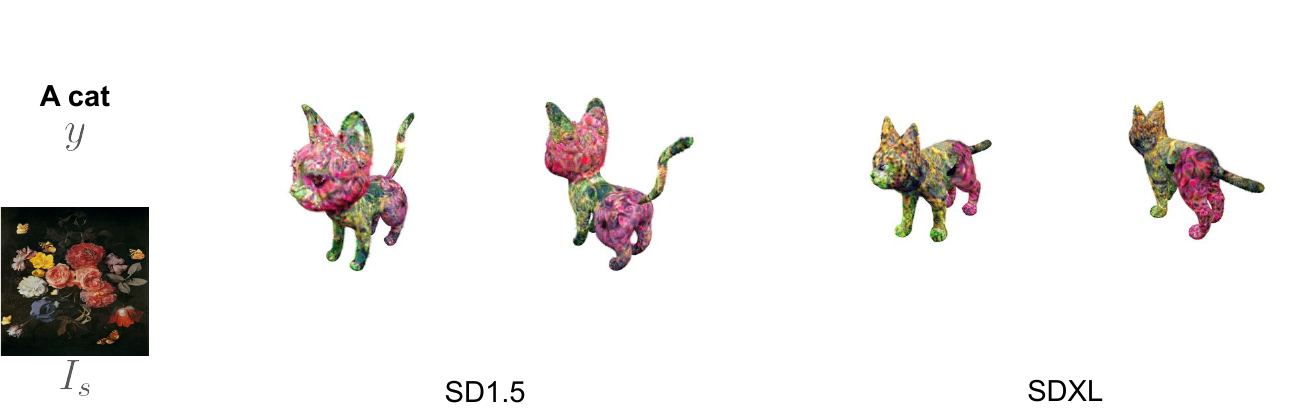}
    \caption{Comparison of stylization results using SD1.5 and SDXL as the backbone diffusion model. SIC3D produces consistent stylization quality across both backbones.}
    \label{fig:sdxl_comparison}
\end{figure}

In addition, while our main results use Stable Diffusion v1.5 for stability and reproducibility, SIC3D is backbone-agnostic. Figure~\ref{fig:sdxl_comparison} provides visual comparisons using SDXL as the underlying diffusion model, demonstrating consistent stylization quality across different backbone choices.

\subsection{More Results}
\label{sec:more_results}
\begin{figure*}[tb]
    \centering
    \includegraphics[width=1\linewidth]{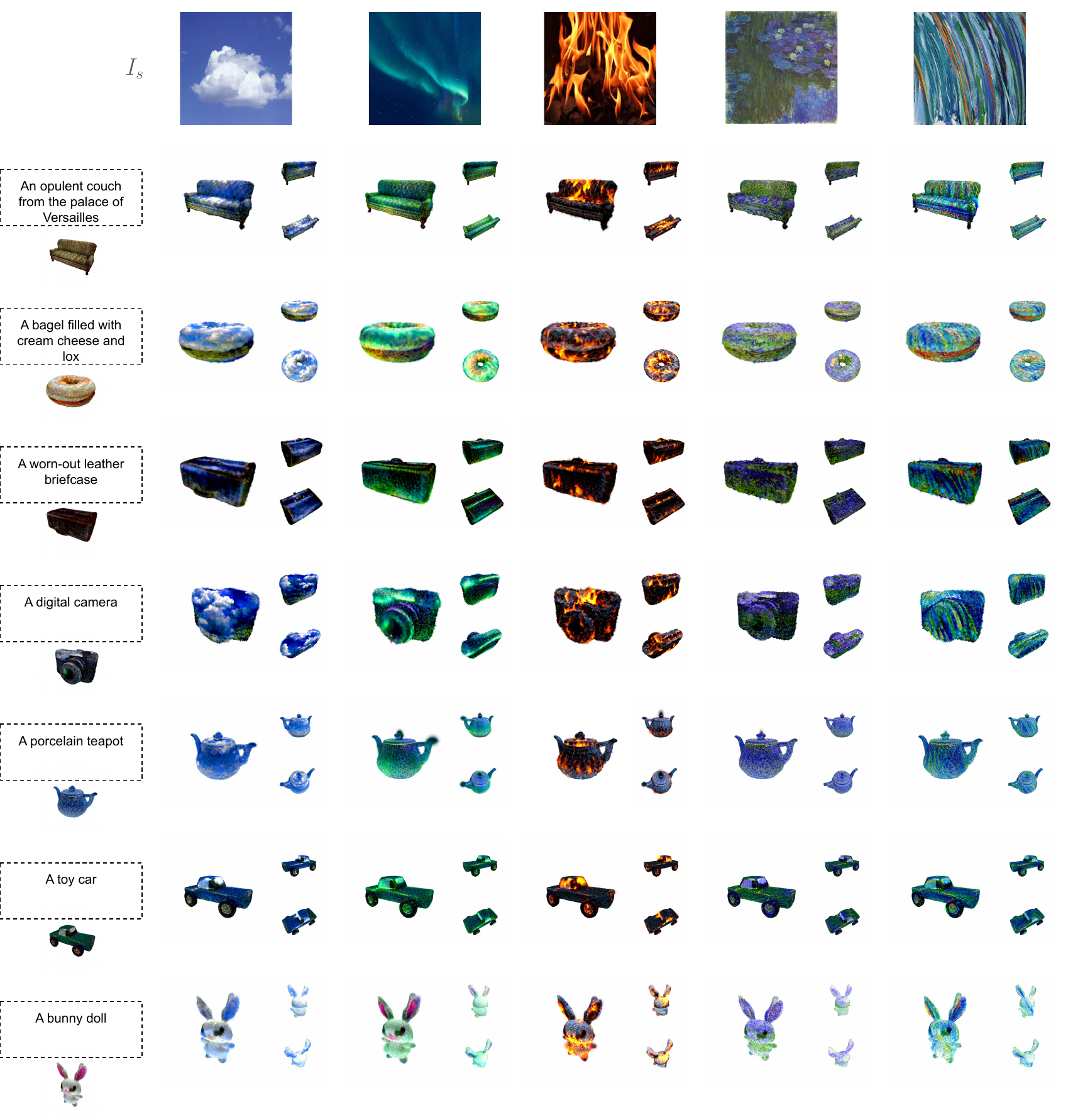}
    \caption{More results from SIC3D}
    \label{fig:more_results}
\end{figure*}

\begin{figure*}[tb]
    \centering
    \includegraphics[width=1\linewidth]{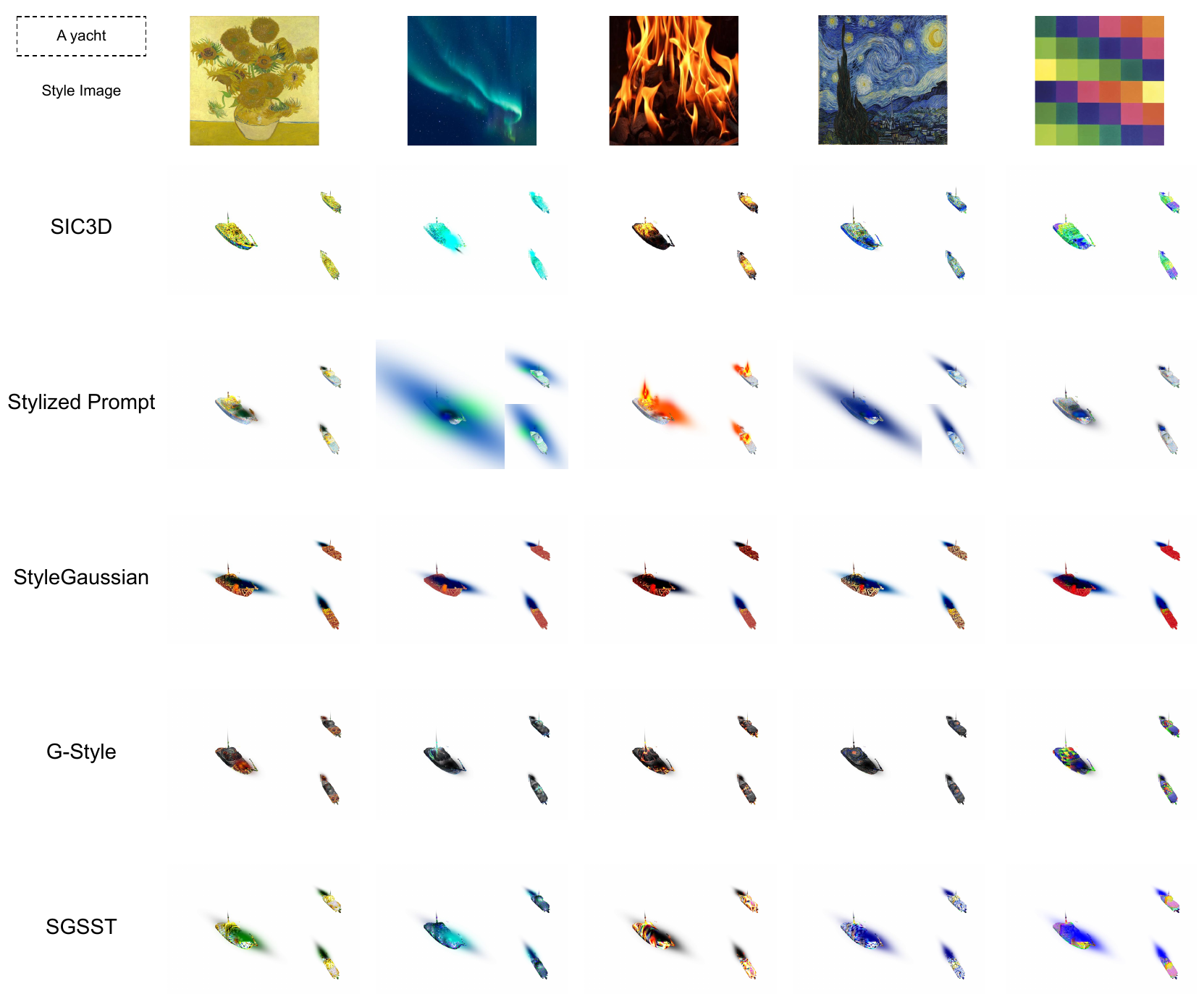}
    \caption{More comparison results between SIC3D and baselines with different style images.}
    \label{fig:more_comparison}
\end{figure*}

We show more comparison results in Figure \ref{fig:more_comparison}, using a single text prompt and different style images. In Figure \ref{fig:more_results}, we present more results with SIC3D using different pairs of text prompt and style image. The IP-Adapter scale used is 0.8 and timestep sampling range is between 0 and 1.

\end{document}